\documentclass[11pt]{article}
\usepackage[final]{acl}
\usepackage{times}
\usepackage{latexsym}
\usepackage[T1]{fontenc}
\usepackage[utf8]{inputenc}
\usepackage{microtype}
\usepackage{booktabs}
\usepackage{multirow}
\usepackage{graphicx}
\usepackage{amsmath}
\usepackage[table]{xcolor}
\usepackage{arydshln}

\definecolor{darkgreen}{HTML}{32CD32}
\definecolor{lightgreen}{HTML}{98FB98}
\definecolor{darkred}{HTML}{B22222}
\definecolor{lightred}{HTML}{FFB6C1}

\title{No Optimal Language Set Exists for Multilingual Instruction Tuning:\\
Insights from a Linguistically-Informed Study}

\author{Gürkan Soykan\thanks{E-mail: gurkan.soykan@wur.nl, Work is done during PhD studies at Koç University.}\textsuperscript{$1$}, Gözde Gül Şahin\textsuperscript{$2,3,4$}
\\[.3em]
\textsuperscript{1} Information Technology Group, Wageningen University and Research, Wageningen, Netherlands \\
\textsuperscript{2} Friedrich-Alexander-Universität Erlangen-Nürnberg, Intelligent Language Systems \\
\textsuperscript{3} Computer Engineering Department, Koç University, Istanbul, Turkey\\
\textsuperscript{4} KUIS AI Lab, Istanbul, Turkey \\
{\url{https://gglab-ku.github.io/}} \\
}

\begin{document}
\maketitle

\begin{abstract}
Multilingual instruction tuning (MIT) is challenged by the curse of multilinguality, data scarcity, and high computational cost. A natural hypothesis is that carefully selecting a linguistically diverse set of languages yields universally better models. We test this systematically by evaluating linguistically-informed selection strategies---based on typological, geographical, semantic, and learned features---against random baselines across three model families (mGPT, mT5-xl, BLOOM) and five multilingual benchmarks. Our key negative finding is that no universal language selection strategy emerges in our fixed-budget setting: performance is strongly task- and model-dependent, and adding more languages beyond a modest threshold triggers degradation consistent with the curse of multilinguality. We discuss implications for MIT data curation and the pitfalls of benchmark-averaged evaluation. All resources are publicly available at \url{https://github.com/GGLAB-KU/ling-informed-mit}.
\end{abstract}

\section{Introduction}

Multilingual language models exhibit uneven cross-lingual performance~\cite{joshi-etal-2020-state}, and instruction tuning with multilingual data has emerged as a practical remedy~\cite{muennighoff2022crosslingual}. However, the curse of multilinguality~\cite{Conneau2019UnsupervisedCR_curse_of_multilinguality}---where per-language performance degrades as more languages compete for model capacity---complicates the pursuit of broad coverage, as do high computational costs and data scarcity.

A natural hypothesis follows: selecting the \textit{right} linguistically diverse subset of languages should yield models that generalize well across languages and tasks. Prior work lends partial support---\citet{chen2023monolingual} and \citet{Shaham2024MultilingualIT} show small multilingual subsets can match or exceed monolingual tuning; \citet{weber-etal-2024-investigating} confirm that parallel multilingual corpora improve cross-lingual instruction-following by up to 9.9\% over monolingual training; and language features have aided cross-lingual transfer in related tasks~\cite{langrank_lin19acl,ustun-etal-2020-udapter}. Yet no prior work has systematically investigated whether linguistically-informed language \textit{selection} yields a universally superior MIT strategy.

We fill this gap. Our central negative finding is that \textbf{no universal language selection strategy was observed in the fixed-budget MIT setting studied here}: the best choice depends on model family and evaluation task, which has important consequences for how the community evaluates and reports MIT results.%{Our central negative finding is that \textbf{there is no single best language selection strategy for MIT}: the optimal choice depends on model family and evaluation task, which has important consequences for how the community evaluates and reports MIT results.}

\section{Experimental Setup}

We use Bactrian-X~\cite{li2023bactrianx}, a parallel multilingual instruction dataset with 3.4M instruction-response pairs across 52 languages. Following the budget-controlled setup of \citet{chen2023monolingual}, we fix total training to 67k instances per experiment (~4,786 per language for 14-language subsets). We select 14 languages per subset using k-means clustering ($k{=}14$) on different feature spaces:
\textbf{GEO} (geographical vectors~\cite{Glottolog}),
\textbf{TYPO} (typological vectors~\cite{littell2017uriel}),
\textbf{SEM} (semantic typology embeddings~\cite{chen-etal-2023-colex2lang}),
\textbf{LEARN} (learned language vectors~\cite{malaviya17emnlp_learned}),
\textbf{FAM} (one language per family),
\textbf{RND} (random, baseline), and
\textbf{ALL} (all 52 languages).
Each is run with 3 random seeds. Table~\ref{tab:subset_langs_main_results} lists the selected languages.

We instruction-tune three model families with LoRA~\cite{hu2021lora}: mGPT-1.3B~\cite{mgpt}, mT5-XL-3.7B~\cite{xue-etal-2021-mt5}, and BLOOM-\{1.7B, 3B, 7B\}~\cite{workshop2023bloom}, evaluated zero-shot on XNLI~\cite{conneau-etal-2018-xnli}, PAWS-X~\cite{yang-etal-2019-paws}, XCOPA~\cite{ponti-etal-2020-xcopa}, XStoryCloze~\cite{xstorycloze}, and XWinograd~\cite{tikhonov2021heads} using the lm-evaluation-harness~\cite{lm_eval_harness}.

\begin{table}[t]
  \centering
  \small
  \caption{Language subsets per strategy and seed (ISO codes).}
  \begin{tabular}{@{}ll@{}}
    \toprule
    \textbf{Subset} & \textbf{Languages} \\ \midrule
    \multirow{3}{*}{FAM}   & az,en,fi,he,ja,ka,vi,ko,ml,mn,my,th,tl,xh \\
                           & ar,en,fi,id,ja,ka,ko,mn,sw,ta,th,tr,vi,zh \\
                           & af,az,et,he,ja,ka,ko,ml,mn,my,th,tl,vi,xh \\ \midrule
    \multirow{3}{*}{GEO}   & af,bn,et,fr,gu,he,hr,id,ja,kk,mn,sw,ta,vi \\
                           & af,bn,et,fr,he,hr,id,ja,ka,kk,mn,ta,ur,vi \\
                           & af,bn,et,fr,he,hr,id,ja,kk,mn,sw,ta,ur,vi \\ \midrule
    \multirow{3}{*}{LEARN} & cs,hi,id,km,ko,lt,lv,my,nl,pt,sl,ta,tl,vi \\
                           & cs,fi,hr,km,ko,lt,lv,my,nl,pt,sl,ta,uk,vi \\
                           & cs,hi,id,ja,lt,lv,nl,pt,sl,sv,ta,tl,tr,vi \\ \midrule
    \multirow{3}{*}{RND}   & cs,gu,hi,id,ko,lv,mk,ml,ps,pt,si,ta,vi,zh \\
                           & cs,en,et,fr,hi,hr,it,ja,ml,mn,mr,ne,pl,pt \\
                           & ar,az,de,en,es,et,he,hi,id,ml,ps,ru,sv,uk \\ \midrule
    \multirow{3}{*}{SEM}   & gl,gu,ka,kk,ko,ml,pl,ps,si,sl,sv,tl,uk,vi \\
                           & en,fr,gl,gu,ka,kk,ko,pl,ps,sl,sv,tl,uk,vi \\
                           & en,gl,gu,ka,kk,ko,ml,pl,ps,sl,sv,tl,uk,vi \\ \midrule
    \multirow{3}{*}{TYPO}  & az,bn,de,et,fa,hi,it,ja,mk,sw,ta,tl,vi,zh \\
                           & ar,az,bn,de,et,fa,it,ja,mk,sw,ta,th,ur,zh \\
                           & ar,az,bn,en,fi,he,hi,it,ja,mk,nl,sw,ta,th \\
    \bottomrule
  \end{tabular}
  \label{tab:subset_langs_main_results}
\end{table}

\section{Results}

\subsection{No Single Strategy Wins Across All Tasks and Models}

Table~\ref{tab:combined_results} presents our main results for BLOOM-7B, mGPT, and mT5-xl. The best-performing strategy varies substantially across models and tasks: GEO leads for BLOOM-7B on XStoryCloze while TYPO achieves the best average for BLOOM-7B overall, FAM is best on average for mT5-xl, and GEO or TYPO lead for mGPT on NLU tasks. Crucially, \textbf{no single subset consistently achieves the highest score across all model-task combinations}.

Improvements over the RND baseline are often not statistically significant. Paired t-tests (three seeds) show significant improvements only in specific model-task combinations---e.g., GEO vs.\ RND for mGPT on XNLI ($p{\leq}0.05$) and XStoryCloze ($p{\leq}0.05$), FAM vs.\ RND for mT5-xl on XCOPA ($p{\leq}0.05$) and XWinograd ($p{\leq}0.05$)---but these do not generalize across all benchmarks or model families. Meanwhile, several linguistically-informed subsets score \emph{significantly lower} than RND in specific configurations (red cells), underscoring that informed selection can also backfire. The effect sizes in Table~\ref{tab:combined_results} contextualise this: Cohen's $d_z$ expresses the mean gain over RND in units of its seed-to-seed variability (0.2/0.5/0.8${=}$small/medium/large) and derives from the same paired t-tests. With $n{=}3$ seeds, $p{\leq}0.05$ requires $|d_z|{\geq}2.5$, so only very large effects can reach significance. Even these are model-specific (GEO: $d_z{=}3.6$ on mGPT vs.\ $0.5$ on BLOOM-7B), and with three seeds individual $d_z$ values are magnitude indicators rather than precise estimates.

For \textbf{PAWS-X}, no strategy produces significant improvement over RND in any model, and the base (non-tuned) model achieves the best PAWS-X score for both BLOOM-7B and mGPT---suggesting MIT may actively harm paraphrase identification performance.

\begin{table*}[t]
    \centering
    \small
    \caption{Main results for BLOOM-7B, mGPT, and mT5-xl. \textbf{Bold}: highest mean per model-task. Color coding vs.\ RND baseline (paired t-test): \colorbox{darkgreen}{dark green} $p{\leq}0.05$, \colorbox{green}{green} $0.05{<}p{\leq}0.1$, \colorbox{lightgreen}{light green} $0.1{<}p{\leq}0.15$ (significantly better); \colorbox{darkred}{dark red}, \colorbox{red}{red}, \colorbox{lightred}{light red} (significantly worse). ``-'' = base model (no instruction tuning). In the Avg.\ column, the parenthesised value is the paired effect size vs.\ RND (Cohen's $d_z$).}
    \scalebox{0.86}{
        \begin{tabular}{l|c|ccccc|c}
            \hline
            \textbf{Model}                                           & \textbf{Subset}
                                                                     & \textbf{XNLI}
                                                                     & \textbf{XCOPA}
                                                                     & \textbf{XStoryCloze}
                                                                     & \textbf{XWinograd}
                                                                     & \textbf{PAWS-X}
                                                                     & \textbf{Avg.} {\tiny($d_z$)}
            \\
            \hline
            \multirow{8}{*}{Bloom-7B}
                                                                     & ALL
                                                                     & \cellcolor{darkred} 40.88 {\tiny $\pm$ 0.25}
                                                                     & \textbf{58.18
            {\tiny $\pm$ 0.54}}                                      & \cellcolor{green} 62.03 {\tiny $\pm$ 0.13}
                                                                     & 73.71 {\tiny
            $\pm$ 0.23}                                              & 47.46 {\tiny $\pm$ 0.82}
                                                                     & 56.45 {\tiny $\pm$ 0.26 ($-$0.1)}
            \\
                                                                     & RND
                                                                     & \textbf{41.63 {\tiny $\pm$ 0.25}}
                                                                     & 57.72 {\tiny
            $\pm$ 0.86}                                              & 61.73 {\tiny $\pm$ 0.30}
                                                                     & 73.91 {\tiny
            $\pm$ 0.31}                                              & 47.36 {\tiny $\pm$ 2.85}
                                                                     & 56.47 {\tiny $\pm$
                    0.69}
            \\
            \cmidrule[0.01pt](r){2-8}
                                                                     & FAM
                                                                     & \cellcolor{red} 41.18 {\tiny $\pm$ 0.67}
                                                                     & 57.95 {\tiny
            $\pm$ 0.19}                                              & 61.49 {\tiny $\pm$ 0.72}
                                                                     & 74.00 {\tiny
            $\pm$ 0.81}                                              & 47.88 {\tiny $\pm$ 2.29}
                                                                     & 56.50 {\tiny $\pm$ 0.56 (0.1)}
            \\
                                                                     & GEO
                                                                     & \cellcolor{red} 41.24 {\tiny $\pm$ 0.66}
                                                                     & 57.97 {\tiny
            $\pm$ 0.39}                                              & \cellcolor{darkgreen} \textbf{62.15 {\tiny $\pm$
            0.31}}                                                   & 73.96 {\tiny
            $\pm$ 1.69}                                              & 47.99 {\tiny $\pm$ 0.85}
                                                                     & 56.66 {\tiny $\pm$ 0.35 (0.5)}
            \\
                                                                     & LEARN
                                                                     & 41.26 {\tiny $\pm$ 0.66}
                                                                     & 58.08 {\tiny
            $\pm$ 0.22}                                              & \cellcolor{green} 61.96 {\tiny $\pm$ 0.06}
                                                                     & \textbf{74.17
            {\tiny $\pm$ 0.29}}                                      & 46.88 {\tiny $\pm$ 1.85}
                                                                     & 56.47 {\tiny $\pm$ 0.21 (0.0)}
            \\
                                                                     & SEM
                                                                     & \cellcolor{red} 41.19 {\tiny $\pm$ 0.54}
                                                                     & 58.03 {\tiny
            $\pm$ 0.66}                                              & 61.44 {\tiny $\pm$ 0.69}
                                                                     &
            \cellcolor{lightgreen} 74.16 {\tiny $\pm$ 0.18}          & 47.35 {\tiny $\pm$ 0.47}
                                                                     & 56.43 {\tiny $\pm$ 0.31 ($-$0.1)}
            \\
                                                                     & TYPO
                                                                     & 41.63 {\tiny $\pm$ 0.80}
                                                                     & 58.09 {\tiny
            $\pm$ 0.54}                                              & 61.94 {\tiny $\pm$ 0.28}
                                                                     & 74.17 {\tiny
            $\pm$ 1.08}                                              & 47.75 {\tiny $\pm$ 0.31}
                                                                     & \textbf{56.71 {\tiny $\pm$ 0.24 (0.8)}}
            \\

                                                                     & -
                                                                     & 41.12
                                                                     & 56.87
                                                                     & 59.30
                                                                     & 73.97
                                                                     & \textbf{49.37}
                                                                     & 56.13
            \\
            \hdashline
            \multirow{8}{*}{mGPT}
                                                                     & ALL
                                                                     & \cellcolor{darkgreen} \textbf{41.23 {\tiny $\pm$
            0.12}}                                                   & 55.82 {\tiny
            $\pm$ 0.14}                                              & 55.98 {\tiny $\pm$ 0.07}
                                                                     &
            \cellcolor{darkred} 60.44 {\tiny $\pm$ 0.30}             & 50.14 {\tiny $\pm$ 0.43}
                                                                     &
            \cellcolor{green} 52.72 {\tiny $\pm$ 0.07 (2.0)}
            \\
                                                                     & RND
                                                                     & 40.71 {\tiny $\pm$ 0.23}
                                                                     & 55.91 {\tiny
            $\pm$ 0.62}                                              & 55.97 {\tiny $\pm$ 0.36}
                                                                     & 60.78 {\tiny
            $\pm$ 0.20}                                              & 48.71 {\tiny $\pm$ 0.99}
                                                                     & 52.42 {\tiny $\pm$
                    0.36}
            \\
            \cmidrule[0.01pt](r){2-8}
                                                                     & FAM
                                                                     & \cellcolor{darkred} 40.27 {\tiny $\pm$ 0.08}
                                                                     & 55.82 {\tiny
            $\pm$ 0.92}                                              & 55.81 {\tiny $\pm$ 0.85}
                                                                     & 60.88 {\tiny
            $\pm$ 1.04}                                              & \cellcolor{lightgreen} 49.69 {\tiny $\pm$ 1.72}
                                                                     & 52.49 {\tiny $\pm$ 0.30 (0.7)}
            \\
                                                                     & GEO
                                                                     & \cellcolor{lightgreen} 41.01 {\tiny $\pm$ 0.29}
                                                                     & 55.73 {\tiny
            $\pm$ 0.56}                                              & \cellcolor{darkgreen} \textbf{56.35 {\tiny $\pm$
            0.13}}                                                   &
            \cellcolor{lightgreen} \textbf{61.05 {\tiny $\pm$ 0.44}} & 49.97 {\tiny $\pm$ 2.64}
                                                                     &
            \cellcolor{darkgreen} \textbf{52.82 {\tiny $\pm$ 0.55 (3.6)}}
            \\
                                                                     & LEARN
                                                                     & 40.64 {\tiny $\pm$ 0.42}
                                                                     & \textbf{55.95
            {\tiny $\pm$ 0.47}}                                      & \cellcolor{green} 56.15 {\tiny $\pm$ 0.27}
                                                                     & 60.66 {\tiny
            $\pm$ 1.03}                                              & 48.92 {\tiny $\pm$ 0.84}
                                                                     & 52.46 {\tiny $\pm$ 0.38 (0.2)}
            \\
                                                                     & SEM
                                                                     & \cellcolor{darkgreen} 40.77 {\tiny $\pm$ 0.21}
                                                                     &
            \cellcolor{lightred} 55.61 {\tiny $\pm$ 0.42}            & 56.14 {\tiny $\pm$ 0.34}
                                                                     &
            \cellcolor{red} 60.57 {\tiny $\pm$ 0.47}                 & 49.17 {\tiny $\pm$ 2.17}
                                                                     & 52.45 {\tiny $\pm$ 0.44 (0.4)}
            \\
                                                                     & TYPO
                                                                     & \cellcolor{green} 40.84 {\tiny $\pm$ 0.31}
                                                                     & 55.93 {\tiny
            $\pm$ 0.57}                                              & \cellcolor{lightgreen} 56.24 {\tiny $\pm$ 0.44}
                                                                     & 60.39 {\tiny
            $\pm$ 0.83}                                              & 49.17 {\tiny $\pm$ 2.06}
                                                                     & 52.51 {\tiny $\pm$ 0.64 (0.7)}
            \\

                                                                     & -
                                                                     & 40.90
                                                                     & 55.04
                                                                     & 54.43
                                                                     & 60.55
                                                                     & \textbf{50.30}
                                                                     & 52.24
            \\
            \hdashline
            \multirow{8}{*}{mT5-xl}
                                                                     & ALL
                                                                     & 36.49 {\tiny $\pm$ 0.69}
                                                                     &
            \cellcolor{green} 53.64 {\tiny $\pm$ 1.01}               & 52.31 {\tiny $\pm$ 0.21}
                                                                     &
            50.39 {\tiny $\pm$ 0.81}                                 & 50.76 {\tiny $\pm$ 1.15}
                                                                     & 48.72 {\tiny $\pm$ 0.31 ($-$0.3)}
            \\
                                                                     & RND
                                                                     & 36.86 {\tiny $\pm$ 0.29}
                                                                     & 53.10 {\tiny
            $\pm$ 0.48}                                              & 52.46 {\tiny $\pm$ 0.52}
                                                                     & 49.88 {\tiny
            $\pm$ 0.75}                                              & 51.74 {\tiny $\pm$ 1.74}
                                                                     & 48.81 {\tiny $\pm$
                    0.55}
            \\
            \cmidrule[0.01pt](r){2-8}
                                                                     & FAM
                                                                     & \cellcolor{darkred} 36.12 {\tiny $\pm$ 0.44}
                                                                     &
            \cellcolor{darkgreen} \textbf{53.72 {\tiny $\pm$ 0.34}}  & 52.59 {\tiny $\pm$ 0.34}
                                                                     &
            \cellcolor{darkgreen} \textbf{51.01 {\tiny $\pm$ 1.69}}  & \textbf{52.28} {\tiny $\pm$ 0.76}
                                                                     &
            \cellcolor{lightgreen} \textbf{49.14 {\tiny $\pm$ 0.19 (1.4)}}
            \\
                                                                     & GEO
                                                                     & \textbf{36.91 {\tiny $\pm$ 0.36}}
                                                                     & 53.23 {\tiny
            $\pm$ 0.23}                                              & 52.61 {\tiny $\pm$ 1.03}
                                                                     &
            \cellcolor{darkgreen} 50.20 {\tiny $\pm$ 0.76}           & 52.11 {\tiny $\pm$ 0.91}
                                                                     & 49.01 {\tiny $\pm$ 0.50 (0.8)}
            \\
                                                                     & LEARN
                                                                     & 36.69 {\tiny $\pm$ 0.55}
                                                                     &
            \cellcolor{lightgreen} 53.48 {\tiny $\pm$ 0.53}          & \textbf{52.62 {\tiny $\pm$ 0.21}}
                                                                     &
            49.65 {\tiny $\pm$ 1.49}                                 & 51.69 {\tiny $\pm$ 1.05}
                                                                     & 48.82 {\tiny $\pm$ 0.13 (0.1)}
            \\
                                                                     & SEM
                                                                     & \cellcolor{darkred} 36.41 {\tiny $\pm$ 0.57}
                                                                     & 53.35 {\tiny
            $\pm$ 0.29}                                              & 52.38 {\tiny $\pm$ 0.33}
                                                                     & 50.62 {\tiny
            $\pm$ 1.89}                                              & 51.63 {\tiny $\pm$ 0.66}
                                                                     & 48.88 {\tiny $\pm$ 0.39 (0.2)}
            \\
                                                                     & TYPO
                                                                     & \cellcolor{darkred} 36.62 {\tiny $\pm$ 0.37}
                                                                     & 53.22 {\tiny
            $\pm$ 0.54}                                              & 52.58 {\tiny $\pm$ 0.34}
                                                                     & 50.09 {\tiny
            $\pm$ 0.86}                                              & 51.40 {\tiny $\pm$ 0.35}
                                                                     & 48.78 {\tiny $\pm$ 0.11 ($-$0.1)}
            \\
                                                                     & -
                                                                     & 33.32
                                                                     & 52.35
                                                                     & 51.36
                                                                     & 50.57
                                                                     & 50.14
                                                                     & 47.55
            \\
            \hline
        \end{tabular}
    }
    \label{tab:combined_results}
\end{table*}

\subsection{More Languages Triggers Curse of Multilinguality}

\begin{figure}[t]
    \centering
    \begin{minipage}[b]{0.49\linewidth}
        \centering
        \includegraphics[width=\linewidth]{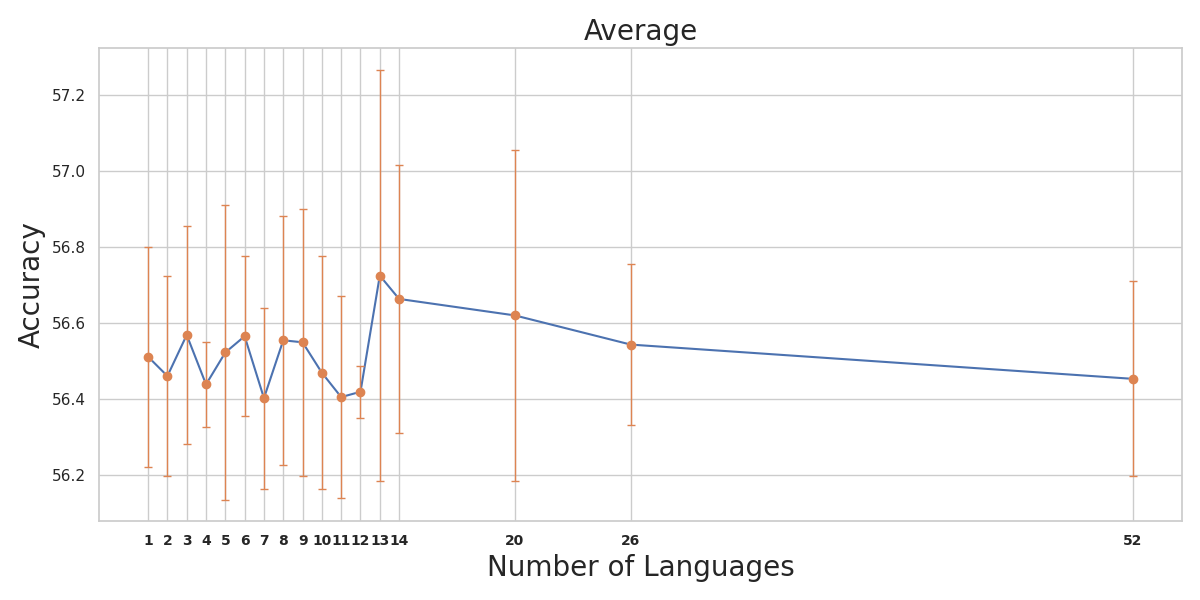}
        \small (a) BLOOM-7B
    \end{minipage}
    \hfill
    \begin{minipage}[b]{0.49\linewidth}
        \centering
        \includegraphics[width=\linewidth]{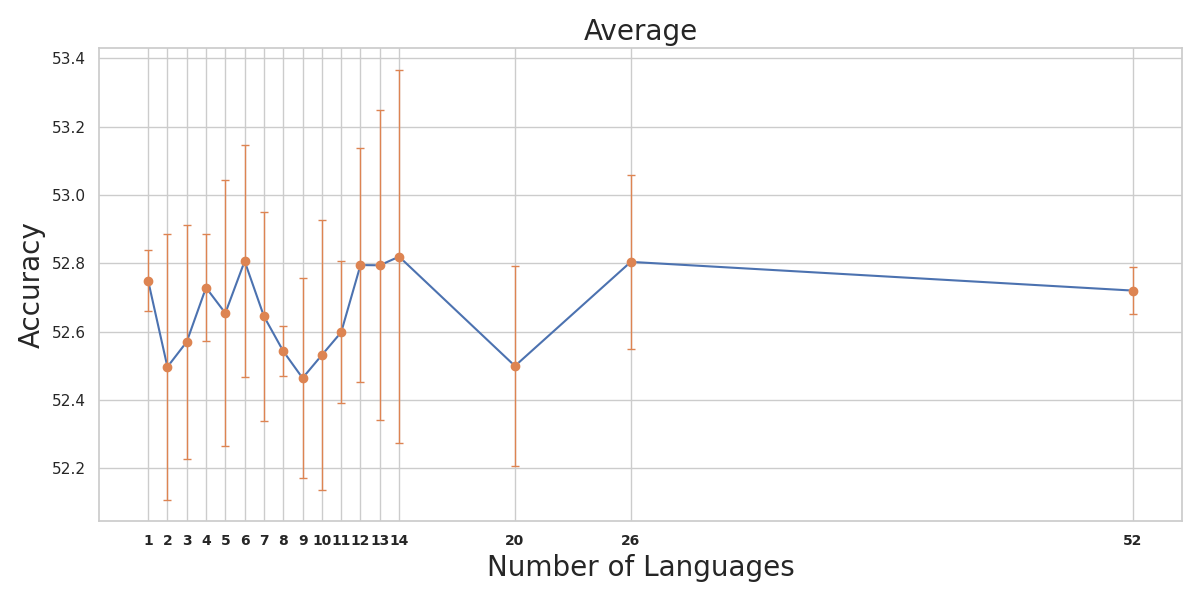}
        \small (b) mGPT
    \end{minipage}
    \caption{Average performance with confidence intervals when varying the number of GEO-selected languages. Performance peaks at 13--14 and declines beyond.}
    \label{fig:geo_varying_langs_mgpt_bloom_merged}
\end{figure}

Figure~\ref{fig:geo_varying_langs_mgpt_bloom_merged} shows the effect of varying language count (1 to 52) using GEO on mGPT and BLOOM-7B. For BLOOM-7B, performance is stable up to 13--14 languages, peaks there, then declines consistently---replicating the curse of multilinguality~\cite{Conneau2019UnsupervisedCR_curse_of_multilinguality} under a fixed computational budget. The ALL setting (52 languages) never achieves best average performance in either model. mGPT shows wider confidence intervals but a similar trend: the 14-language default performs marginally better than larger configurations.

\subsection{Cross-lingual Generalization to Unseen Languages is Model-Dependent}

\begin{figure}[t]
    \centering
    \includegraphics[width=\linewidth]{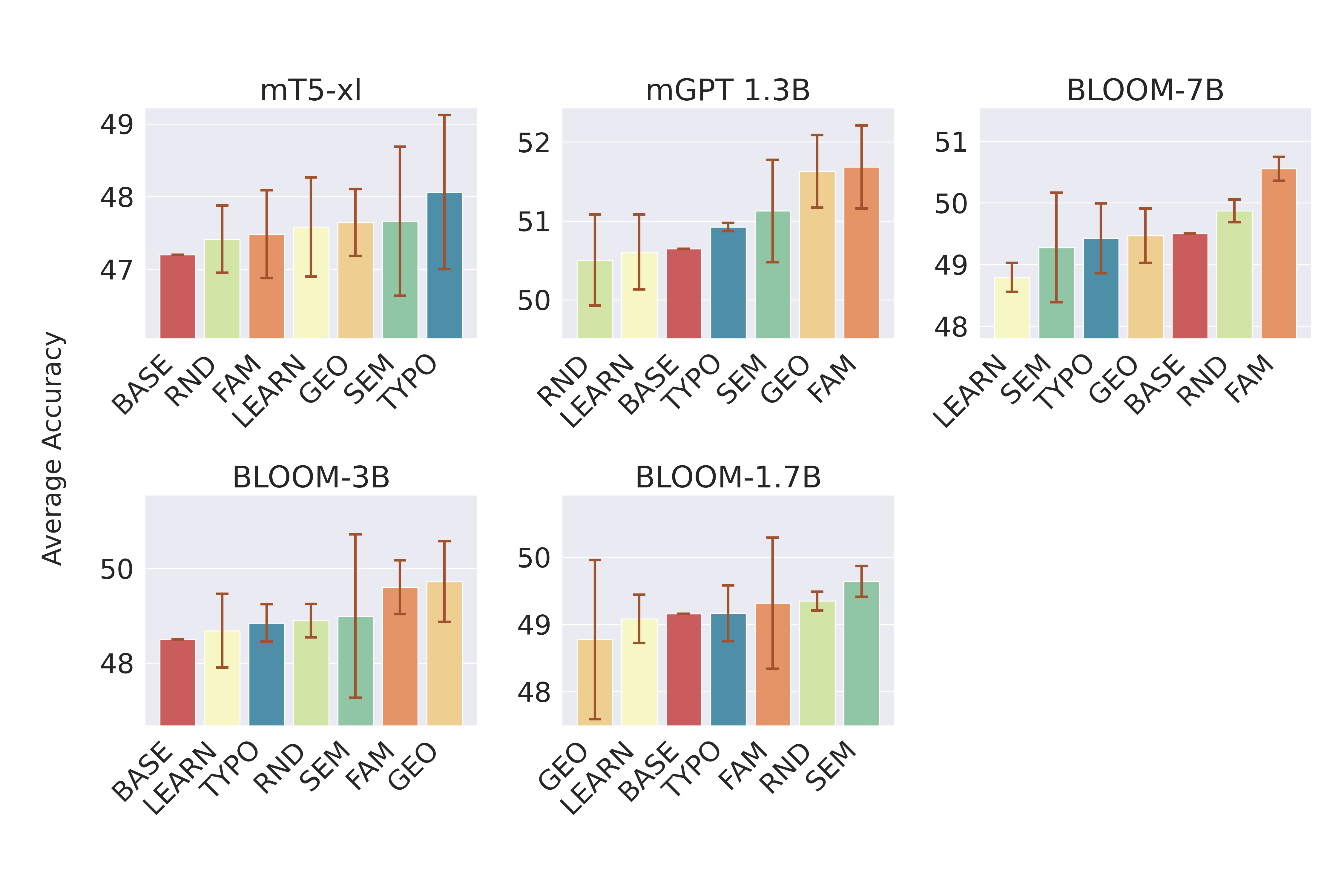}
    \caption{Average performance on unseen languages (not part of any instruction-tuning subset).}
    \label{fig:not_seen_perf}
\end{figure}

Figure~\ref{fig:not_seen_perf} shows model performance on \textit{unseen} languages (those in evaluation benchmarks but not in any training subset). While linguistically-informed subsets generally improve over the non-tuned base model for mGPT, mT5-xl, and BLOOM-3B, results diverge for larger BLOOM models. For BLOOM-7B, only two of the five subsets improve over the base. This reveals a non-trivial interaction between pretraining corpus composition and subset selection: a strategy effective for one model's pretraining distribution may be harmful for another's.

\subsection{Model Size Amplifies Differences---But Not Uniformly}

\begin{figure}[t]
    \centering
    \includegraphics[width=\linewidth]{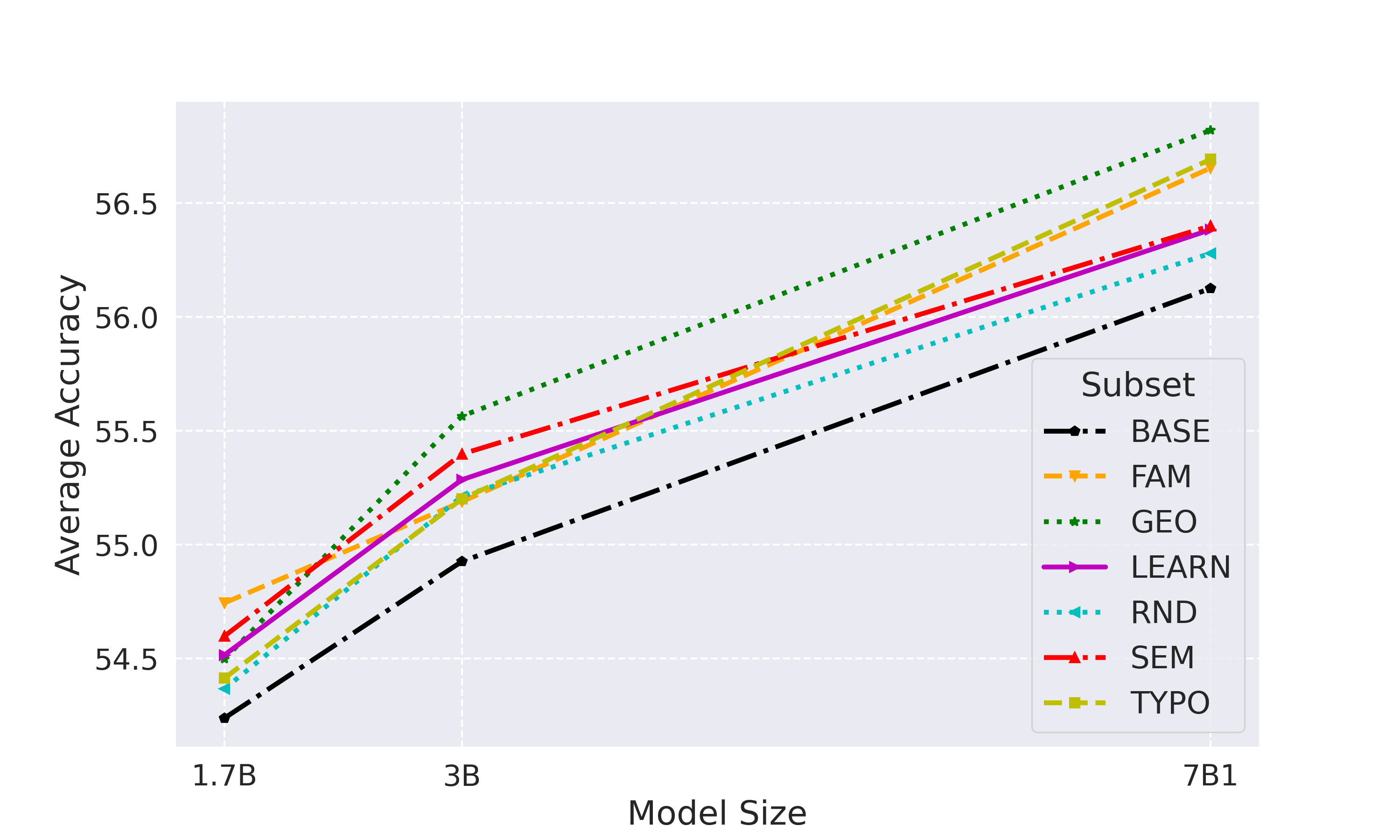}
    \caption{Average zero-shot performance across BLOOM model sizes by language subset. Strategy rankings shift with scale.}
    \label{fig:bloom_scaling_effect}
\end{figure}

Figure~\ref{fig:bloom_scaling_effect} shows how language subset performance evolves across BLOOM-1.7B, 3B, and 7B. While all instruction-tuned models surpass the base, the random baseline consistently ranks lowest among tuned models. GEO excels at 3B and 7B, whereas FAM starts strong at 1.7B but diminishes at scale. TYPO improves progressively with scale, reaching second place at 7B. Importantly, \textbf{the ranking of strategies changes across model sizes}: there is no strategy that is uniformly best as scale increases.

\subsection{Monolingual Tuning Remains Competitive}

\begin{figure}[t]
    \centering
    \includegraphics[width=\linewidth]{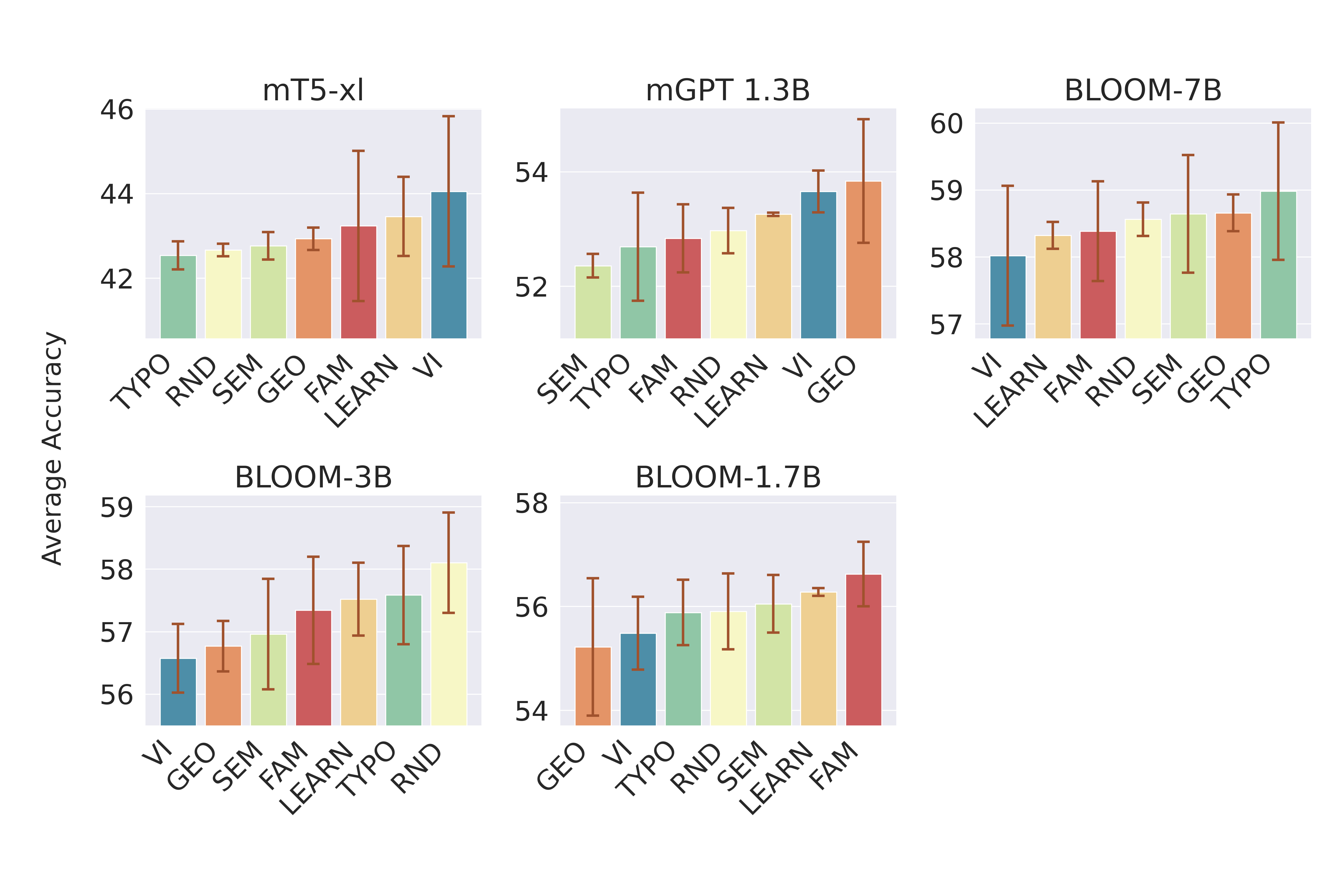}
    \caption{Monolingual (VI) vs.\ multilingual instruction tuning for Vietnamese on XNLI and XCOPA. Monolingual is competitive for mT5 and mGPT but not BLOOM.}
    \label{fig:vi_mono_multi_perf}
\end{figure}

In a Vietnamese case study (Figure~\ref{fig:vi_mono_multi_perf}), monolingual tuning (67k Vietnamese examples) ranks first for mT5 and second for mGPT on average across XNLI and XCOPA---with no statistically significant difference from the best multilingual subset (GEO) for mGPT. For BLOOM, the monolingual case consistently underperforms. This partially undermines the assumption that multilingual subsets are always preferable, and shows that the right choice is again model-dependent.

\subsection{Training--Evaluation Language Overlap}
\label{sec:overlap}

Table~\ref{tab:xnli_overlap} shows, for XNLI, how much overlap exists between each model's pretraining languages (M), each training subset (S), and the evaluation languages (T). BLOOM stands out: its pretraining corpus covers only 8 of the 15 XNLI languages, and it has the smallest three-way overlaps of all models. This matches Table~\ref{tab:combined_results}, where most informed subsets score significantly below RND on XNLI for BLOOM-7B. 

mGPT instead covers 14 of the 15 XNLI languages, and the same informed subsets score significantly higher there. We find that overlap alone does not decide the outcomes, however: GEO improves XStoryCloze significantly on both BLOOM-7B and mGPT while sharing only 2 of its 11 evaluation languages. Since benchmark language sets themselves barely intersect (XNLI covers 15 languages, XWinograd 6), no single subset can align with every task.

\begin{table}[t]
  \centering
  \small
  \caption{Language intersections for XNLI: counts of shared languages between the model's pretraining corpus (M), the instruction tuning language subset (S), and the task's evaluation languages (T). Bloom = BLOOM-7B.}
  \begin{tabular}{@{}llcccc@{}}
    \toprule
    \textbf{Model} & \textbf{Subset} & \textbf{M-S-T} & \textbf{S-T} & \textbf{M-T} & \textbf{M-S} \\ \midrule
    mT5-xl & ALL   & 13/15 & 13/15 & 15/15 & 50/52 \\
    mGPT   & ALL   & 12/15 & 13/15 & 14/15 & 39/52 \\
    mT5-xl & RND   &  9/15 &  9/15 & 15/15 & 31/32 \\
    mT5-xl & TYPO  &  9/15 &  9/15 & 15/15 & 20/21 \\
    Bloom  & ALL   &  8/15 & 13/15 &  8/15 & 18/46 \\
    mGPT   & RND   &  8/15 &  9/15 & 14/15 & 24/32 \\
    mGPT   & TYPO  &  8/15 &  9/15 & 14/15 & 18/21 \\
    mT5-xl & FAM   &  7/15 &  7/15 & 15/15 & 21/22 \\
    Bloom  & RND   &  6/15 &  9/15 &  8/15 & 13/32 \\
    Bloom  & TYPO  &  6/15 &  9/15 &  8/15 &  8/21 \\
    mGPT   & FAM   &  6/15 &  7/15 & 14/15 & 20/22 \\
    Bloom  & FAM   &  4/15 &  7/15 &  8/15 &  8/22 \\
    mT5-xl & GEO   &  4/15 &  4/15 & 15/15 & 15/16 \\
    mGPT   & GEO   &  4/15 &  4/15 & 14/15 & 14/16 \\
    Bloom  & GEO   &  4/15 &  4/15 &  8/15 &  8/16 \\
    mT5-xl & LEARN &  3/15 &  3/15 & 15/15 & 18/20 \\
    mT5-xl & SEM   &  3/15 &  3/15 & 15/15 & 15/16 \\
    mGPT   & LEARN &  3/15 &  3/15 & 14/15 & 15/20 \\
    mGPT   & SEM   &  3/15 &  3/15 & 14/15 & 11/16 \\
    Bloom  & SEM   &  3/15 &  3/15 &  8/15 &  5/16 \\
    Bloom  & LEARN &  2/15 &  3/15 &  8/15 &  5/20 \\
    \bottomrule
  \end{tabular}
  \label{tab:xnli_overlap}
\end{table}

\section{Discussion and Implications}

\paragraph{Why no optimal set exists.} Three compounding factors explain the absence of a universal winner. (1)~\textit{Pretraining corpus mismatch}: languages that complement one model's pretraining gaps differ across mGPT (61 languages), BLOOM (59, different sources), and mT5 (101). We quantify this: for BLOOM, a language's accuracy closely tracks its share of the pretraining corpus, both before tuning (pooled Spearman $\rho{=}0.55$--$0.63$, $p{<}10^{-4}$) and after ($\rho{=}0.58$--$0.65$). For mT5, whose sampling flattens language proportions, the correlation vanishes ($\rho{=}{-}0.04$, n.s.). Yet a subset's aggregate pretraining share does not predict tuned performance (n.s.\ in four of five models). Hence, pretraining explains where models perform well, but not which subset to tune on.
(2)~\textit{Task--language alignment} (\S\ref{sec:overlap}): high subset overlap with XNLI's 15 languages does not imply overlap with XWinograd's 6 or XCOPA's 11, making universal optimality structurally impossible. (3)~\textit{Seed sensitivity}: three-seed confidence intervals are wide enough that strategy differences often fall within noise. These factors are not unique to our setting: \citet{shimabucoro-etal-2025-posttrain} reach the same conclusion at larger scale (up to 35B parameters), finding CLT dynamics vary jointly with task type, model size, and training regime.

\paragraph{Methodological warning.} Reporting single aggregate scores across tasks and models in MIT can be deeply misleading. A strategy that appears superior in aggregate may underperform significantly on specific tasks: FAM achieves the best average for mT5-xl but scores worst on XNLI; GEO leads for mGPT overall but shows negative effects on BLOOM for the same task. Strikingly, the base non-tuned model scores best on PAWS-X for both BLOOM-7B and mGPT, suggesting that MIT can \textit{harm} performance on certain task types. We urge the community to always report per-task and per-model breakdowns with confidence intervals, and to avoid over-generalizing from a limited set of benchmarks or model families. Additionally, the zero-shot multiple-choice format of standard evaluation harnesses~\cite{lm_eval_harness} has been critiqued for misaligning with real-world model use~\cite{lyu-etal-2024-beyond}, which may further explain cross-task variability.

\paragraph{Practical takeaway.} Despite the absence of a universal optimum, linguistically-informed selection---particularly GEO and TYPO---\textit{tends} to be a safer default than random selection, especially at larger model scales. For practitioners without resources to run full ablations, GEO-based selection via URIEL features~\cite{littell2017uriel} provides a compute-free heuristic that beats random on average. Alternatively, representation-space approaches such as LangGPS~\cite{ye-etal-2025-langgps} offer a data-driven signal when model access is available. The key practical insight for dataset curators is to target 10--15 linguistically diverse languages rather than maximising language count: more languages is not better under a fixed computational budget.

\section{Conclusion}

We set out to find the optimal set of languages for multilingual instruction tuning. In our fixed-budget setting, we did not find one. Linguistically-informed language selection generally outperforms random selection on average, but no tested strategy is consistently best across tasks, model families, and evaluation settings.%{We set out to find the optimal set of languages for multilingual instruction tuning. We did not find one. Linguistically-informed language selection generally outperforms random selection on average, but no single strategy is consistently best across tasks, model families, and evaluation settings.} 
The curse of multilinguality reasserts itself beyond approximately 14 languages under a fixed computational budget; cross-lingual generalization to unseen languages is highly model-dependent; monolingual tuning can be competitive with multilingual approaches; and strategy rankings shift with model scale. Taken together, these negative findings caution against simple universal prescriptions for MIT data curation. They also highlight the need for task- and model-specific evaluation with proper statistical reporting---a standard the field has not yet widely adopted.

\section*{Limitations}

Several limitations bear on the generalisability of our findings. First, we rely exclusively on LoRA~\cite{hu2021lora} as the PEFT method; other fine-tuning approaches may yield different sensitivity to language subset choice. Second, Bactrian-X~\cite{li2023bactrianx} consists largely of automatically translated English instructions, potentially introducing translation artefacts that confound language-specific effects.
To gain more insights, we scored each language's translation quality as the mean LaBSE~\cite{Feng2020LanguageagnosticBS} similarity between 500 English source instructions and their translations. We find that quality does not significantly correlate with per-language performance (Spearman $\rho{=}0.09$--$0.24$, n.s., vs.\ $0.58$ for pretraining share over the same languages). Furthermore, the selection strategies barely differ in the mean quality of the languages they pick ($0.862$--$0.877$). Translation quality is therefore unlikely to explain our negative result, although LaBSE is itself an imperfect proxy.

In addition, all experiments are capped at 7B parameters due to resource constraints; behaviour at larger scales remains unknown. Fourth, only three random seeds per configuration limit statistical power, leaving confidence intervals wider than desirable---a key reason many strategy comparisons fail to reach significance. Fifth, language selection is constrained to the 52 Bactrian-X languages; a broader or different pool could yield different subsets. 

Finally, the zero-shot multiple-choice evaluation format may not reflect instruction-following quality in open-ended generation settings~\cite{lyu-etal-2024-beyond,lm_eval_Biderman2024LessonsFT}, which is the more relevant real-world use case for instruction-tuned models. However, the models studied here generate outputs of too low quality for a meaningful open-ended evaluation, as we observed in preliminary LLM-as-a-judge experiments.

\section*{Acknowledgments}
  This work has been supported by the Scientific and Technological Research Council of Türkiye~(TÜBİTAK) as part of
  the project ``Automatic Learning of Procedural Language from Natural Language Instructions for Intelligent
  Assistance'' with the number 121C132. We also gratefully acknowledge KUIS AI Lab for providing computational support.

\clearpage
\bibliography{refs}

@inproceedings{ustun-etal-2020-udapter,
  title     = {{UD}apter: Language Adaptation for Truly {U}niversal {D}ependency Parsing},
  author    = {{\"U}st{\"u}n, Ahmet  and
               Bisazza, Arianna  and
               Bouma, Gosse  and
               van Noord, Gertjan},
  editor    = {Webber, Bonnie  and
               Cohn, Trevor  and
               He, Yulan  and
               Liu, Yang},
  booktitle = {Proceedings of the 2020 Conference on Empirical Methods in Natural Language Processing (EMNLP)},
  month     = nov,
  year      = {2020},
  address   = {Online},
  publisher = {Association for Computational Linguistics},
  url       = {https://aclanthology.org/2020.emnlp-main.180},
  doi       = {10.18653/v1/2020.emnlp-main.180},
  pages     = {2302--2315}
}

@inproceedings{chen-etal-2023-colex2lang,
  title     = {{C}olex2{L}ang: Language Embeddings from Semantic Typology},
  author    = {Chen, Yiyi  and
               Biswas, Russa  and
               Bjerva, Johannes},
  editor    = {Alum{\"a}e, Tanel  and
               Fishel, Mark},
  booktitle = {Proceedings of the 24th Nordic Conference on Computational Linguistics (NoDaLiDa)},
  month     = may,
  year      = {2023},
  address   = {T{\'o}rshavn, Faroe Islands},
  publisher = {University of Tartu Library},
  url       = {https://aclanthology.org/2023.nodalida-1.67},
  pages     = {673--684}
}

@article{chen2023monolingual,
  title   = {Monolingual or multilingual instruction tuning: Which makes a better alpaca},
  author  = {Chen, Pinzhen and Ji, Shaoxiong and Bogoychev, Nikolay and Haddow, Barry and Heafield, Kenneth},
  journal = {arXiv preprint arXiv:2309.08958},
  year    = {2023}
}

@article{Shaham2024MultilingualIT,
  title   = {Multilingual Instruction Tuning With Just a Pinch of Multilinguality},
  author  = {Uri Shaham and Jonathan Herzig and Roee Aharoni and Idan Szpektor and Reut Tsarfaty and Matan Eyal},
  journal = {ArXiv},
  year    = {2024},
  volume  = {abs/2401.01854},
  url     = {https://api.semanticscholar.org/CorpusID:266741580}
}

@misc{li2023bactrianx,
  title         = {Bactrian-X : A Multilingual Replicable Instruction-Following Model with Low-Rank Adaptation},
  author        = {Haonan Li and Fajri Koto and Minghao Wu and Alham Fikri Aji and Timothy Baldwin},
  year          = {2023},
  eprint        = {2305.15011},
  archiveprefix = {arXiv},
  primaryclass  = {cs.CL}
}

@inproceedings{littell2017uriel,
  title     = {Uriel and lang2vec: Representing languages as typological, geographical, and phylogenetic vectors},
  author    = {Littell, Patrick and Mortensen, David R and Lin, Ke and Kairis, Katherine and Turner, Carlisle and Levin, Lori},
  booktitle = {Proceedings of the 15th Conference of the European Chapter of the Association for Computational Linguistics: Volume 2, Short Papers},
  volume    = {2},
  pages     = {8--14},
  year      = {2017}
}

@inproceedings{malaviya17emnlp_learned,
  title     = {Learning Language Representations for Typology Prediction},
  author    = {Malaviya, Chaitanya and Neubig, Graham and Littell, Patrick},
  booktitle = {Conference on Empirical Methods in Natural Language Processing (EMNLP)},
  address   = {Copenhagen, Denmark},
  month     = {September},
  year      = {2017}
}

@online{Glottolog,
  author       = {Harald Hammarström and Robert Forkel and Martin Haspelmath and Sebastian Bank},
  title        = {Glottolog database 4.8},
  year         = {2023},
  url          = {https://glottolog.org},
  organization = {Max Planck Institute for Evolutionary Anthropology},
  address      = {Leipzig},
  note         = {Available online at https://glottolog.org}
}

@misc{mgpt,
  doi       = {10.48550/ARXIV.2204.07580},
  url       = {https://arxiv.org/abs/2204.07580},
  author    = {Shliazhko, Oleh and Fenogenova, Alena and Tikhonova, Maria and Mikhailov, Vladislav and Kozlova, Anastasia and Shavrina, Tatiana},
  title     = {mGPT: Few-Shot Learners Go Multilingual},
  publisher = {arXiv},
  year      = {2022},
  copyright = {Creative Commons Attribution 4.0 International}
}

@article{workshop2023bloom,
  author     = {Teven Le Scao and
                Angela Fan and
                Christopher Akiki and
                Ellie Pavlick and
                Suzana Ilic and
                Daniel Hesslow and
                Roman Castagn{\'{e}} and
                Alexandra Sasha Luccioni and
                Fran{\c{c}}ois Yvon and
                Matthias Gall{\'{e}} and
                Jonathan Tow and
                Alexander M. Rush and
                Stella Biderman and
                Albert Webson and
                Pawan Sasanka Ammanamanchi and
                Thomas Wang and
                Beno{\^{\i}}t Sagot and
                Niklas Muennighoff and
                Albert Villanova del Moral and
                Olatunji Ruwase and
                Rachel Bawden and
                Stas Bekman and
                Angelina McMillan{-}Major and
                Iz Beltagy and
                Huu Nguyen and
                Lucile Saulnier and
                Samson Tan and
                Pedro Ortiz Suarez and
                Victor Sanh and
                Hugo Lauren{\c{c}}on and
                Yacine Jernite and
                Julien Launay and
                Margaret Mitchell and
                Colin Raffel and
                Aaron Gokaslan and
                Adi Simhi and
                Aitor Soroa and
                Alham Fikri Aji and
                Amit Alfassy and
                Anna Rogers and
                Ariel Kreisberg Nitzav and
                Canwen Xu and
                Chenghao Mou and
                Chris Emezue and
                Christopher Klamm and
                Colin Leong and
                Daniel van Strien and
                David Ifeoluwa Adelani and
                et al.},
  title      = {{BLOOM:} {A} 176B-Parameter Open-Access Multilingual Language Model},
  journal    = {CoRR},
  volume     = {abs/2211.05100},
  year       = {2022},
  url        = {https://doi.org/10.48550/arXiv.2211.05100},
  doi        = {10.48550/ARXIV.2211.05100},
  eprinttype = {arXiv},
  eprint     = {2211.05100},
  bibsource  = {dblp computer science bibliography, https://dblp.org}
}

@inproceedings{xue-etal-2021-mt5,
  title     = {m{T}5: A Massively Multilingual Pre-trained Text-to-Text Transformer},
  author    = {Xue, Linting  and
               Constant, Noah  and
               Roberts, Adam  and
               Kale, Mihir  and
               Al-Rfou, Rami  and
               Siddhant, Aditya  and
               Barua, Aditya  and
               Raffel, Colin},
  editor    = {Toutanova, Kristina  and
               Rumshisky, Anna  and
               Zettlemoyer, Luke  and
               Hakkani-Tur, Dilek  and
               Beltagy, Iz  and
               Bethard, Steven  and
               Cotterell, Ryan  and
               Chakraborty, Tanmoy  and
               Zhou, Yichao},
  booktitle = {Proceedings of the 2021 Conference of the North American Chapter of the Association for Computational Linguistics: Human Language Technologies},
  month     = jun,
  year      = {2021},
  address   = {Online},
  publisher = {Association for Computational Linguistics},
  url       = {https://aclanthology.org/2021.naacl-main.41},
  doi       = {10.18653/v1/2021.naacl-main.41},
  pages     = {483--498}
}

@misc{lm_eval_harness,
  author    = {Gao, Leo and Tow, Jonathan and Abbasi, Baber and Biderman, Stella and Black, Sid and DiPofi, Anthony and Foster, Charles and Golding, Laurence and Hsu, Jeffrey and Le Noac'h, Alain and Li, Haonan and McDonell, Kyle and Muennighoff, Niklas and Ociepa, Chris and Phang, Jason and Reynolds, Laria and Schoelkopf, Hailey and Skowron, Aviya and Sutawika, Lintang and Tang, Eric and Thite, Anish and Wang, Ben and Wang, Kevin and Zou, Andy},
  title     = {A framework for few-shot language model evaluation},
  month     = 12,
  year      = 2023,
  publisher = {Zenodo},
  version   = {v0.4.0},
  doi       = {10.5281/zenodo.10256836},
  url       = {https://zenodo.org/records/10256836}
}

@inproceedings{ponti-etal-2020-xcopa,
  title     = {{XCOPA}: A Multilingual Dataset for Causal Commonsense Reasoning},
  author    = {Ponti, Edoardo Maria  and
               Glava{\v{s}}, Goran  and
               Majewska, Olga  and
               Liu, Qianchu  and
               Vuli{\'c}, Ivan  and
               Korhonen, Anna},
  editor    = {Webber, Bonnie  and
               Cohn, Trevor  and
               He, Yulan  and
               Liu, Yang},
  booktitle = {Proceedings of the 2020 Conference on Empirical Methods in Natural Language Processing (EMNLP)},
  month     = nov,
  year      = {2020},
  address   = {Online},
  publisher = {Association for Computational Linguistics},
  url       = {https://aclanthology.org/2020.emnlp-main.185},
  doi       = {10.18653/v1/2020.emnlp-main.185},
  pages     = {2362--2376}
}

@article{xstorycloze,
  author     = {Xi Victoria Lin and
                Todor Mihaylov and
                Mikel Artetxe and
                Tianlu Wang and
                Shuohui Chen and
                Daniel Simig and
                Myle Ott and
                Naman Goyal and
                Shruti Bhosale and
                Jingfei Du and
                Ramakanth Pasunuru and
                Sam Shleifer and
                Punit Singh Koura and
                Vishrav Chaudhary and
                Brian O'Horo and
                Jeff Wang and
                Luke Zettlemoyer and
                Zornitsa Kozareva and
                Mona T. Diab and
                Veselin Stoyanov and
                Xian Li},
  title      = {Few-shot Learning with Multilingual Language Models},
  journal    = {CoRR},
  volume     = {abs/2112.10668},
  year       = {2021},
  url        = {https://arxiv.org/abs/2112.10668},
  eprinttype = {arXiv},
  eprint     = {2112.10668},
  bibsource  = {dblp computer science bibliography, https://dblp.org}
}

@misc{muennighoff2022crosslingual,
  title         = {Crosslingual Generalization through Multitask Finetuning},
  author        = {Niklas Muennighoff and Thomas Wang and Lintang Sutawika and Adam Roberts and Stella Biderman and Teven Le Scao and M Saiful Bari and Sheng Shen and Zheng-Xin Yong and Hailey Schoelkopf and Xiangru Tang and Dragomir Radev and Alham Fikri Aji and Khalid Almubarak and Samuel Albanie and Zaid Alyafeai and Albert Webson and Edward Raff and Colin Raffel},
  year          = {2022},
  eprint        = {2211.01786},
  archiveprefix = {arXiv},
  primaryclass  = {cs.CL}
}

@misc{tikhonov2021heads,
  title         = {It's All in the Heads: Using Attention Heads as a Baseline for Cross-Lingual Transfer in Commonsense Reasoning},
  author        = {Alexey Tikhonov and Max Ryabinin},
  year          = {2021},
  eprint        = {2106.12066},
  archiveprefix = {arXiv},
  primaryclass  = {cs.CL}
}

@inproceedings{conneau-etal-2018-xnli,
  title     = {{XNLI}: Evaluating Cross-lingual Sentence Representations},
  author    = {Conneau, Alexis  and
               Rinott, Ruty  and
               Lample, Guillaume  and
               Williams, Adina  and
               Bowman, Samuel  and
               Schwenk, Holger  and
               Stoyanov, Veselin},
  editor    = {Riloff, Ellen  and
               Chiang, David  and
               Hockenmaier, Julia  and
               Tsujii, Jun{'}ichi},
  booktitle = {Proceedings of the 2018 Conference on Empirical Methods in Natural Language Processing},
  month     = oct # {-} # nov,
  year      = {2018},
  address   = {Brussels, Belgium},
  publisher = {Association for Computational Linguistics},
  url       = {https://aclanthology.org/D18-1269},
  doi       = {10.18653/v1/D18-1269},
  pages     = {2475--2485}
}

@inproceedings{yang-etal-2019-paws,
  title     = {{PAWS}-{X}: A Cross-lingual Adversarial Dataset for Paraphrase Identification},
  author    = {Yang, Yinfei  and
               Zhang, Yuan  and
               Tar, Chris  and
               Baldridge, Jason},
  booktitle = {Proceedings of the 2019 Conference on Empirical Methods in Natural Language Processing and the 9th International Joint Conference on Natural Language Processing (EMNLP-IJCNLP)},
  month     = nov,
  year      = {2019},
  address   = {Hong Kong, China},
  publisher = {Association for Computational Linguistics},
  url       = {https://aclanthology.org/D19-1382},
  doi       = {10.18653/v1/D19-1382},
  pages     = {3687--3692}
}

@inproceedings{lyu-etal-2024-beyond,
  title     = {Beyond Probabilities: Unveiling the Misalignment in Evaluating Large Language Models},
  author    = {Lyu, Chenyang  and
               Wu, Minghao  and
               Aji, Alham},
  editor    = {Li, Sha  and
               Li, Manling  and
               Zhang, Michael JQ  and
               Choi, Eunsol  and
               Geva, Mor  and
               Hase, Peter  and
               Ji, Heng},
  booktitle = {Proceedings of the 1st Workshop on Towards Knowledgeable Language Models (KnowLLM 2024)},
  month     = aug,
  year      = {2024},
  address   = {Bangkok, Thailand},
  publisher = {Association for Computational Linguistics},
  url       = {https://aclanthology.org/2024.knowllm-1.10},
  pages     = {109--131}
}

@article{lm_eval_Biderman2024LessonsFT,
  title   = {Lessons from the Trenches on Reproducible Evaluation of Language Models},
  author  = {Stella Biderman and Hailey Schoelkopf and Lintang Sutawika and Leo Gao and Jonathan Tow and Baber Abbasi and Alham Fikri Aji and Pawan Sasanka Ammanamanchi and Sid Black and Jordan Clive and Anthony DiPofi and Julen Etxaniz and Benjamin Fattori and Jessica Zosa Forde and Charles Foster and Mimansa Jaiswal and Wilson Y. Lee and Haonan Li and Charles Lovering and Niklas Muennighoff and Ellie Pavlick and Jason Phang and Aviya Skowron and Samson Tan and Xiangru Tang and Kevin A. Wang and Genta Indra Winata and Franccois Yvon and Andy Zou},
  journal = {ArXiv},
  year    = {2024},
  volume  = {abs/2405.14782},
  url     = {https://api.semanticscholar.org/CorpusID:269982020}
}

@article{hu2021lora,
  title   = {Lora: Low-rank adaptation of large language models},
  author  = {Hu, Edward J and Shen, Yelong and Wallis, Phillip and Allen-Zhu, Zeyuan and Li, Yuanzhi and Wang, Shean and Wang, Lu and Chen, Weizhu},
  journal = {arXiv preprint arXiv:2106.09685},
  year    = {2021}
}

@inproceedings{langrank_lin19acl,
  title     = {Choosing Transfer Languages for Cross-Lingual Learning},
  author    = {Yu-Hsiang Lin and Chian-Yu Chen and Jean Lee and Zirui Li and Yuyan Zhang and Mengzhou Xia and Shruti Rijhwani and Junxian He and Zhisong Zhang and Xuezhe Ma and Antonios Anastasopoulos and Patrick Littell and Graham Neubig},
  booktitle = {The 57th Annual Meeting of the Association for Computational Linguistics (ACL)},
  address   = {Florence, Italy},
  month     = {July},
  year      = {2019}
}

@inproceedings{Conneau2019UnsupervisedCR_curse_of_multilinguality,
  title     = {Unsupervised Cross-lingual Representation Learning at Scale},
  author    = {Alexis Conneau and Kartikay Khandelwal and Naman Goyal and Vishrav Chaudhary and Guillaume Wenzek and Francisco Guzm{\'a}n and Edouard Grave and Myle Ott and Luke Zettlemoyer and Veselin Stoyanov},
  booktitle = {Annual Meeting of the Association for Computational Linguistics},
  year      = {2019},
  url       = {https://api.semanticscholar.org/CorpusID:207880568}
}

@inproceedings{joshi-etal-2020-state,
  title     = {The State and Fate of Linguistic Diversity and Inclusion in the {NLP} World},
  author    = {Joshi, Pratik  and
               Santy, Sebastin  and
               Budhiraja, Amar  and
               Bali, Kalika  and
               Choudhury, Monojit},
  editor    = {Jurafsky, Dan  and
               Chai, Joyce  and
               Schluter, Natalie  and
               Tetreault, Joel},
  booktitle = {Proceedings of the 58th Annual Meeting of the Association for Computational Linguistics},
  month     = jul,
  year      = {2020},
  address   = {Online},
  publisher = {Association for Computational Linguistics},
  url       = {https://aclanthology.org/2020.acl-main.560},
  doi       = {10.18653/v1/2020.acl-main.560},
  pages     = {6282--6293}
}

@inproceedings{weber-etal-2024-investigating,
  title     = {Investigating Multilingual Instruction-Tuning: Do Polyglot Models Demand for Multilingual Instructions?},
  author    = {Weber, Alexander Arno and Thellmann, Klaudia and Ebert, Jan and Flores-Herr, Nicolas and Lehmann, Jens and Fromm, Michael and Ali, Mehdi},
  booktitle = {Proceedings of the 2024 Conference on Empirical Methods in Natural Language Processing},
  year      = {2024},
  pages     = {20829--20855},
  address   = {Miami, Florida, USA},
  publisher = {Association for Computational Linguistics},
  doi       = {10.18653/v1/2024.emnlp-main.1159}
}

@article{shimabucoro-etal-2025-posttrain,
  title   = {A Post-trainer's Guide to Multilingual Training Data: Uncovering Cross-lingual Transfer Dynamics},
  author  = {Shimabucoro, Lu{\'\i}sa and {\"U}st{\"u}n, Ahmet and Fadaee, Marzieh and Ruder, Sebastian},
  journal = {arXiv preprint arXiv:2504.16677},
  year    = {2025}
}

@inproceedings{ye-etal-2025-langgps,
  title     = {{LangGPS}: Language Separability Guided Data Pre-Selection for Joint Multilingual Instruction Tuning},
  author    = {Ye, Yangfan and Feng, Xiaocheng and Feng, Xiachong and Huang, Lei and Ma, Weitao and Hong, Qichen and Lu, Yunfei and Tang, Duyu and Tu, Dandan and Qin, Bing},
  booktitle = {Proceedings of the AAAI Conference on Artificial Intelligence},
  year      = {2026}
}

@inproceedings{Feng2020LanguageagnosticBS,
  title={{Language-agnostic BERT Sentence Embedding}},
  author={Fangxiaoyu Feng and Yinfei Yang and Daniel Matthew Cer and N. Arivazhagan and Wei Wang},
  booktitle={Annual Meeting of the Association for Computational Linguistics},
  year={2020}
}

\end{document}